\documentclass[letterpaper, 10 pt, conference]{ieeeconf}  

\IEEEoverridecommandlockouts
\IEEEoverridecommandlockouts                              

\usepackage{soul}
\usepackage{cite}
\usepackage{amsmath,amssymb,amsfonts}
\usepackage{graphicx}
\usepackage{textcomp}
\usepackage{algpseudocode} 
\usepackage{mathrsfs}
\def\BibTeX{{\rm B\kern-.05em{\sc i\kern-.025em b}\kern-.08em
    T\kern-.1667em\lower.7ex\hbox{E}\kern-.125emX}}
\usepackage{subcaption}

\usepackage{booktabs}
\usepackage{pifont}
\newcommand{\cmark}{\ding{51}}%
\newcommand{\xmark}{\ding{55}}%

\usepackage{pifont}
\usepackage[table]{xcolor}

\begin{document}

\title{\LARGE \bf Edge Computing and its Application in Robotics: A Survey}

\author{Nazish Tahir \and Ramviyas Parasuraman
\thanks{The authors are with the School of Computing, University of Georgia, Athens, GA 30602, USA. Authors email: {\tt\small \{nazish.tahir,ramviyas\}@uga.edu}}}

\maketitle

\begin{abstract}
The Edge computing paradigm has gained prominence in both academic and industry circles in recent years. By implementing edge computing facilities and services in robotics, it becomes a key enabler in the deployment of artificial intelligence applications to robots. \textcolor{black}{Time-sensitive robotics applications benefit from the reduced latency, mobility, and location awareness provided by the edge computing paradigm, which enables real-time data processing and intelligence at the network's edge. While the advantages of integrating edge computing into robotics are numerous, there has been no recent survey that comprehensively examines these benefits. This paper aims to bridge that gap by highlighting important work in the domain of edge robotics, examining recent advancements, and offering deeper insight into the challenges and motivations behind both current and emerging solutions. In particular, this article provides a comprehensive evaluation of recent developments in edge robotics, with an emphasis on fundamental applications, providing in-depth analysis of the key motivations, challenges, and future directions in this rapidly evolving domain. It also explores the importance of edge computing in real-world robotics scenarios where rapid response times are critical. Finally, the paper outlines various open research challenges in the field of edge robotics.}
\end{abstract}

\section{Introduction}
Edge computing is emerging as a new research hotspot in the computing landscape \cite{sun2016edgeiot},\cite{alrawais2017fog}, \cite{hassan2018role}. It is a novel computing paradigm that performs computation at the network's edge. Edge computing differs from traditional cloud computing concepts in that it allows computation to take place closer to the source of data. It provides the same services as cloud servers, but in close proximity, which facilitates quicker processing and hence faster execution time \cite{khan2019edge}.  Zha et al. \cite{zhao2018edge} defined the concept of edge computing. ``Edge computing is a computing model that unifies resources that are close to the user in geographical distance or network distance to provide computing, storage, and network services for applications."

Edge computing, in essence, enables the migration of the cloud's network, computing, and storage capabilities to the network's edge and the provision of intelligent services at the edge to meet the high computational demands of the data optimization and application intelligence of the fast-paced Information Technology (IT) industry. \textcolor{black}{It facilitates the execution of} application intelligence on the network with low latency and high bandwidth. It expands the cloud network by enhancing computation, storage, and resource capabilities \textcolor{black}{at} the network's edge, close to the source of data, with the goal of meeting critical needs such as real-time servicing, application intelligence, security, and privacy, as well as the network's criteria for low latency and high bandwidth \cite{maskeliunas2019review}.

\begin{figure}[t]
    \centering
    \begin{center}
    \includegraphics[width=.45\textwidth]{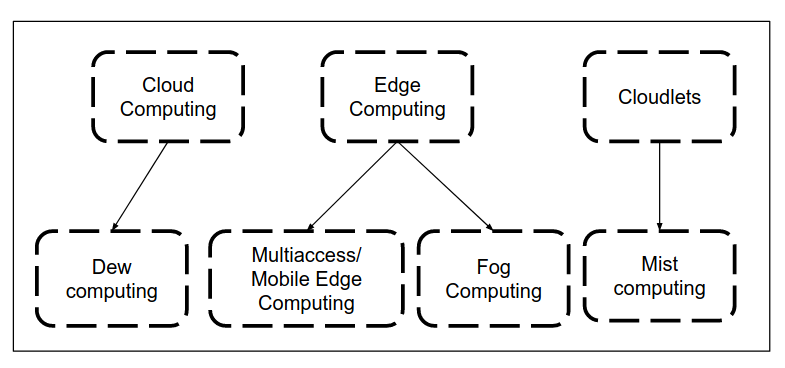}
    \caption{Taxonomy of existing computational domains. \\ \textit{Note: Fog computing is often used interchangeably with Edge computing in literature and is classified accordingly here.}}
    \label{fig:overview}
    \vspace{-1mm}
\end{center}
\end{figure}

A brief taxonomy of existing computational domains related to cloud computing concepts is presented in Fig.~\ref{fig:overview}.

The promising features of edge computing are \textit{fast data processing and analysis}, leading to quicker response time and speed. Since edge devices are closer to the data source, they eliminate the need for intermediate data pre-processing and \textcolor{black}{enable} a faster transmission process. One of the most crucial aspects of edge computing is its \textit{proximity to users}, which allows for the provision of better data-intelligent services due to increased digital data transfer over networks \textcolor{black}{across} shorter distances, \textit{minimizing latency}. Furthermore, unlike cloud computation, which entails total offloading of data for processing and a centralized processing approach, it allows for \textit{distributed and parallel processing}, saving time and ensuring data security \cite{cao2020overview}.  It offers \textit{low-cost, low bandwidth cost, and energy consumption} \cite{chen2004studying} which makes it an ideal alternative to cloud computing. 

Outsourcing all the required tasks to the cloud data center might place a significant strain on the network and trigger network congestion and network delays, particularly for large data volume exchange. Because of such network delays, the cloud is inadequate for addressing real-time, low-latency, and high-quality-of-service applications. The use of edge reduces not only the stress on network bandwidth but also the entire energy consumption of local devices associated with data transmission and processing, \textcolor{black}{thereby} increasing computing efficiency significantly. 
\begin{table}[h]
  \centering
  \caption{List of acronyms and their description}
  \label{tab1}
  \begin{tabular}{@{}p{1.8cm}llll@{}}
    \toprule
    Acronym & Description\\
    \midrule
    ETSI & European Telecommunications Standards Institute \\
    MEC & Mobile Edge Computing  \\
    CPS & Cyber Physical Systems\\
    SLAM & Simultaneous Localization \& Mapping \\
    AWS & Amazon Web Services \\
    ROS & Robot Operating System \\
    MRS & Multi Robot Systems\\
    RAN & Radio Access Network \\
    FCN & Fog Computing Nodes \\
    OS & Operating System \\
    VM & Virtual Machines \\
    GPU & Graphics Processing Unit \\
    NPU & Neural Network Processor Unit \\
    \textcolor{black}{FPGA} & Field Programmable Gate Array \\
    TPU & Tensor Processing Unit \\
    APU & Accelerated Processing Unit \\
    UAV & Unmanned Aerial Vehicle \\
    LSTM & Long Short-Term Memory \\
    MRS & Multi-Robot System \\
    \bottomrule
  \end{tabular}
\end{table}

\subsection{Applications of Edge Computing in Robotics}
In recent years, there has been a significant emphasis in the industry on developing smart Cyber-Physical Systems (CPSs) that are deployed in healthcare, transportation, and agriculture sectors to conduct complex engineering processes with less direct human intervention and \textcolor{black}{improved} cost and performance. Robots are a crucial component of this CPSs system, which \textcolor{black}{rely on} automated tasks in the sense-compute-act cycle \cite{spatharakis2022resource}. 

Although a robot is equipped with sufficient computing resources to perform small-scale operations, it is heavily reliant on remote resources to perform large-scale computations, as the robot's onboard computational capacity may be insufficient to perform real-time data processing on large chunks of data \cite{tahir2022analog}. Offloading enables robot operators to access data from anywhere at any time by providing global storage \textcolor{black}{and processing capabilities}. Since the concept of CPS is heavily reliant on the timely delivery of suitable data with the node placement to the relevant computing entities, robots that move resource-intensive tasks to remote resources require efficient solutions with \textcolor{black}{minimal} latency. 

To accomplish this objective, the concept of cloud computing has been applied to multirobot systems, known as cloud robotics \cite{hu2012cloud}, \cite{wan2016cloud}. \textcolor{black}{In this paradigm, the cloud provides} computational resources such as virtual machines or containers, as well as resources from both local and remote data centers, allowing scalable and vast data processing \cite{rahman2019energy}. However, because Cloud servers are located in remote places, most of which are multi-hop distance away, connection time and data transfer latency might be categorized as a key drawback of Cloud robotics, particularly for time-sensitive applications \cite{tardioli2017pound}. 

\begin{figure}[htbp]
  \centering
  \includegraphics[width=0.8\linewidth]{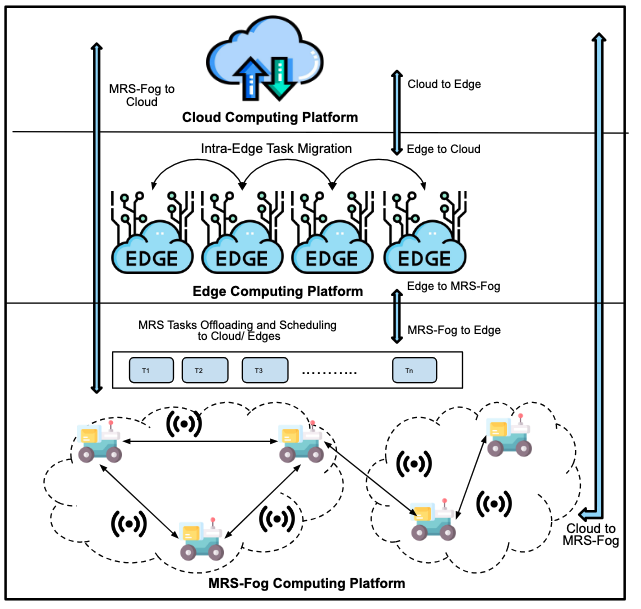} 
  \caption{\textcolor{black}{Overview of robot applications supported by edge, cloud, and fog (multi-robot collaboration) infrastructures.}}
  \label{fig:edge_robotics}
\end{figure}

Edge computing \textcolor{black}{offers a promising solution to this limitation by providing} cloud-like infrastructure at the network's edge, allowing mobile robots to access computational and storage resources. Edge computing capitalizes on the necessity for real-time task execution in close proximity to these smart systems, such as image processing and path planning (Fig. \ref{fig:edge_robotics}). In the context of CPS, edge computing enables robots to conduct a variety of robotic applications with remote support. It provides abstraction to classifying robotic activities based on computing needs, \textcolor{black}{and offers} distributed resources to analyze large amounts of data with minimal communication latency \cite{afrin2021resource}. It also boasts the robots' competency in knowledge exchange by leveraging local edge devices rather than distant cloud servers. Moreover, it contributes to lowering the bandwidth required for data transfer to the cloud. 

\textcolor{black}{\subsection{Survey Scope}}
\textcolor{black}{Several studies published in reputable journals have surveyed the use of edge computing in robotics, with a primary focus on specific experimental prototypes and system architectures. These works have been instrumental in highlighting the limitations of cloud computing in addressing the operational challenges of robotic systems. For example, M. Groshev et al. \cite{groshev2023edge} provide a valuable contribution by analyzing the current landscape of edge computing in robotics and experimentally evaluating an end-to-end robotic system based on existing solutions. Their study identifies key challenges related to system operation and network limitations, as well as the need for improved orchestration mechanisms. However, while their work offers important insights into specific applications and experimental results, it lacks a comprehensive review of the broader literature and does not provide a detailed classification of existing solutions in edge robotics.}

\textcolor{black}{
Other works, such as \cite{haidegger2019robotics}, take a broader societal and cross-domain perspective, focusing on taxonomy, standardization, and human-machine interaction, rather than specifically addressing edge computing or architectural evaluations in multi-robot systems. While they touch on certain aspects relevant to Edge Robotics, they do not provide a comprehensive or focused study of the field. }

\textcolor{black}{
There has been other efforts in comprehensively designing a survey study of edge computing in robotics in specific fields like agriculture \cite{10550846}, industry 5.0 with one section focusing on industrial robots \cite{10462492} and healthcare service robotics \cite{wan2020cognitive}.}

\begin{figure}[htbp]
  \centering
  \includegraphics[width=0.99\linewidth]{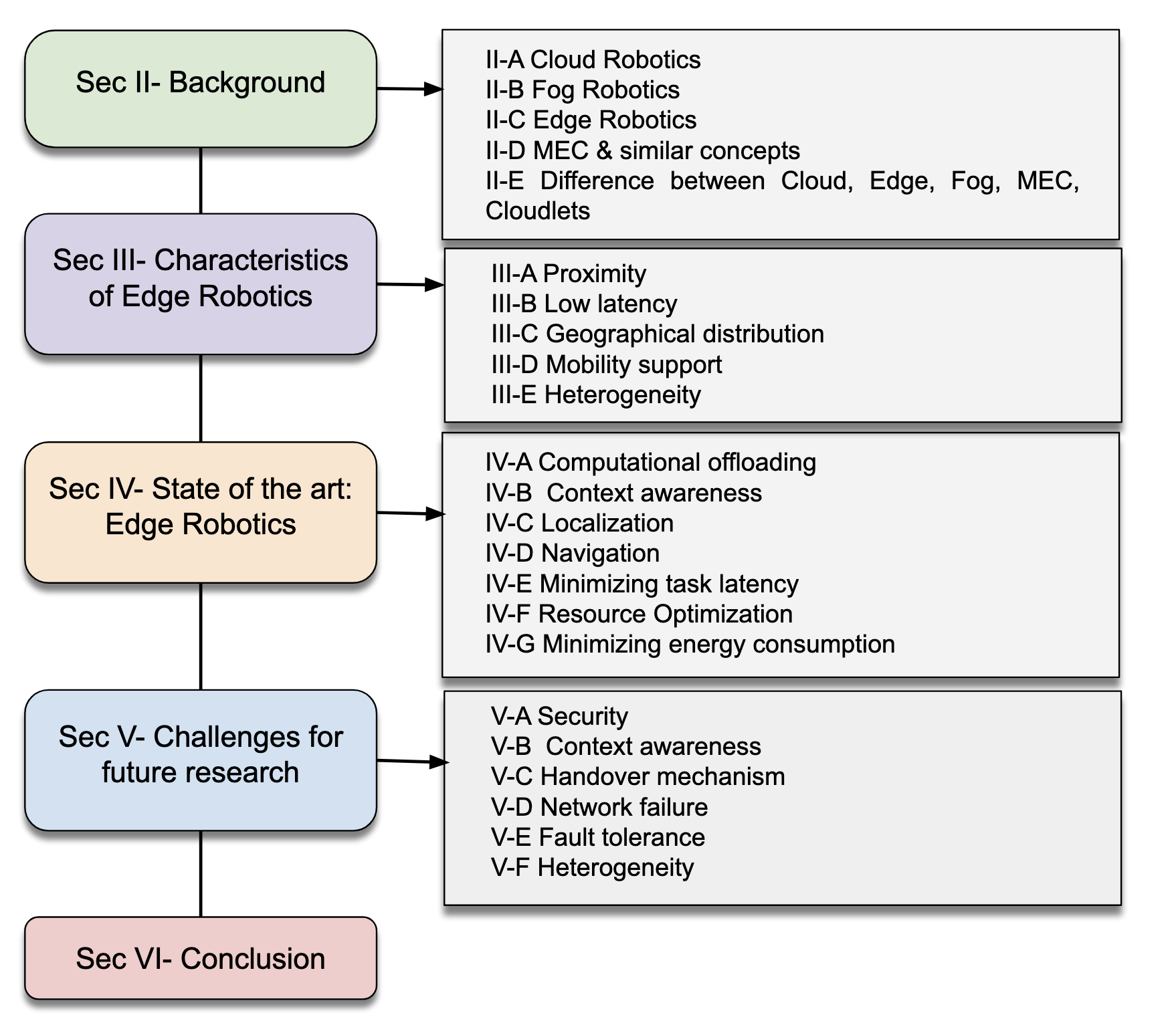} 
  \caption{\textcolor{black}{Table of content and survey organization }}
  \label{fig:content}
\end{figure}
\textcolor{black}{Our work distinguishes itself by presenting a comprehensive survey of Edge Robotics, encompassing a structured classification based on application domains, comparative analysis, and an in-depth discussion of open research challenges. We begin by establishing the necessary background to distinguish among various distributed computing paradigms, positioning Edge Robotics within this broader context. The survey then investigates the core characteristics of Edge Robotics, reviews state-of-the-art applications across a range of robotic tasks, and concludes with a discussion of key challenges and potential future directions. Thus, the main contributions of this paper are summarized as follows:}
\textcolor{black}{
\begin{itemize}
    \item A classification of the diverse computing paradigms relevant to Edge Robotics
    \item A survey of the current landscape and core aspects of Edge Robotics
    \item An identification and discussion of key research challenges and future directions
\end{itemize}
We conducted an in-depth literature review using three major academic databases—IEEE Xplore, ScienceDirect, and the ACM Digital Library—to investigate current and emerging research on edge robotics. By employing search terms such as edge computing in robotics, multi-robot edge systems, multi-edge servers in mutirobot systems, and edge robotics, we identified a substantial body of relevant publications spanning from 2015 to 2025. This review provided valuable insights into the evolution of edge robotics, highlighting both historical developments and the current state of the technology. }

\textcolor{black}{\subsection{Article Organization}}
The remainder of the article is organized as follows: Section \ref{sec:background} introduces the fundamental principles of cloud robotics, edge robotics, fog robotics, MEC, and other similar topics. Section \ref{sec:edge_robotics} delves into the properties of Edge Robotics as well as the reason for its adoption. Section \ref{sec:stateoftheart} investigates the state-of-the-art in Edge robotics and presents a comparative evaluation of research work that shows the benefits and drawbacks of existing frameworks. Section \ref{sec:challenges} delves into the open research challenges of Edge Robotics implementation. Section \ref{sec:conclusion} brings the paper to a close. For a better understanding, the organization of this article is also depicted in Fig~\ref{fig:content}.

\section{Background} 
\label{sec:background}

\subsection{Cloud Robotics}
The term ``cloud robotics" was first coined by Kuffner in 2010 \cite{kuffner2010cloud}. It represents any robot or automation system that \textcolor{black}{utilizes} cloud infrastructure for data or code execution - that is, a system in which all sensing, computing, and memory are not \textcolor{black}{confined} into a single independent unit.  Cloud-enabled robots \textcolor{black}{can perform} complex tasks considerably more efficiently by leveraging massively parallel computation, grid computing capabilities, \textcolor{black}{and} enhanced storage capacity \cite{wan2016cloud}. 

Cloud computing \textcolor{black}{offers} dynamic and on-demand computing resources with centralized storage that can be accessed from anywhere at any time \textcolor{black}{enabling} remote robot control and monitoring \cite{hu2012cloud}. Cloud robotics fosters collaboration among geographically dispersed robots by providing access to \textcolor{black}{extensive datasets, including} global maps for localization, object models for manipulation, high computational machine learning techniques, and \textcolor{black}{shared code repositories}. \cite{kehoe2015survey}. There are several cloud robotics system architectures available, like DAvinCi \cite{arumugam2010davinci}, which is a software framework based on a Hadoop cluster with Robot Operating System (ROS) as a messaging framework, which can be efficiently utilized in applications like FastSLAM \cite{montemerlo2007fastslam}.

Another popular open-source cloud robotics platform called Rapyuta \cite{mohanarajah2014rapyuta}, driven by cloud robotics, enables the robots to offload computational tasks while offering them access to the RoboEarth knowledge database. Similarly, C\^2TAM \cite{riazuelo2014c2tam} is a cloud robotics framework especially designed for cooperative tracking and mapping to solve SLAM problems. Amazon's AWS RobotMaker \cite{sartoni2022aws} has recently emerged as a cloud-based simulation tool that allows developers to perform simulations without \textcolor{black}{the need to manage underlying} infrastructure.

Cloud robotics applications \textcolor{black}{span a wide range of domains including} localization, navigation, grasping and manipulation, perception and computer vision, service robots, social and medical robots, manufacturing, and human-robot interaction. The current research problems in cloud robotics include establishing effective resource and task allocation schemes via the Cloud, minimizing communication delays associated with network and data processing, and \textcolor{black}{enhancing privacy and security for} cloud-based robotic applications \cite{saha2018comprehensive}.

\subsection{Fog Robotics}
In order to overcome the challenges associated with cloud robotics, Cisco introduced the concept of fog computing \cite{bonomi2012fog}. Based on the same concept as edge robotics, fog robotics (FR) can be defined as an architecture that deploys small platforms placed at the network edge to offer storage, network functions, and control with decentralized computing closer to robots. The fog robotics architecture also involves a cloud system in addition to a Fog Robot Server (FRS) \cite{gudi2017fog}. Any multi-robot system (MRS) requests the fog layer for data access and, upon availability, can efficiently utilize the resources available at the fog without querying the cloud \cite{latif2023communication}.

The benefits of using fog robotics for robotic applications are closely similar to the ones offered by edge robotics i.e., offering remote computational offloading, remote monitoring and control of the robots, security services, communication services, storage and caching services, offering global synchronization and collaboration within MRSs, mobility support, navigation and scalability. Fog Robotics helps assist household robotics by allowing models trained on the cloud to be regularly pushed to a smart home gateway rather than directly downloaded to individual robots. In other words, it offers a local model repository cache to robots so that they may adapt models or run a shared inference service for local robots to support those with restricted onboard resources.

Different architectures are supported by the fog-based MRS system \cite{8358727}. They have been classified as \textit{Nanoscale MRS} designed to efficiently service nanoscale robots, \textit{Body Area MRS} which comprises of an MRS system deployed on a single mobile body like a human or a mobile structure like wearable robots, that require wireless connectivity with a single fog node, \textit{Local Area MRS} that is deployed in a limited geographic area, allowing robots to connect to each other through a wireless connected fog layer for better performance, reliability and scalability.

There are various issues associated with fog robotics that may be investigated in the literature. The data stored at the fog layer should be maintained depending on the local movement of the robots; therefore, effective data maintenance and storage are required. There is a handover delay during data transfer between robots and fog servers that should be examined and regulated. Robust security protocols must be in place to prevent malicious robot users from accessing fog-based systems. Additionally, the status of a robot's battery must be taken into account when processing tasks at the fog so that the robot can complete the entire task execution, including data transfer from the fog layer. 

\subsection{Edge Robotics}
Edge robotics \textcolor{black}{aims to} address the issues associated with cloud \textcolor{black}{and} fog robotics by extending the storage and processing capacity of a large number of robots linked to a network, offering an intermediary layer between the robots and the cloud.
\textcolor{black}{Unlike fog robotics}, edge robotics can handle computational data distributively at the edge devices without the need to offload it to the cloud. Fog computing nodes (FCNs) handle the task by computing and storing the data from mobile robots locally before handover to the Cloud, while edge nodes may solve the data query solely on the edge layer. The existence of network edges reduces the computational load of mobile robots by allowing them to handle some of the requests addressed to the cloud locally \textcolor{black}{- those} that do not require cloud intervention. This minimizes delay in resolving queries and enables real-time handling of either a subset of requests or the entire dataset in parallel. 

\subsection{MEC and similar concepts}
\textit{Mist Computing} is another term that \textcolor{black}{originates from} the traditional edge computing domain. Mist computing \textcolor{black}{involves} the use of complementary embedded peripheral devices to preprocess or filter sensory input before sending it to the fog or cloud level. Complex sensors and actuators, such as cameras, 3D scanners, and laser range finders, are equipped with a microcontroller unit or field-programmable gate array that may execute certain functions; nevertheless, their processing capabilities are restricted. Mist computing is the initial computing point in the robot-fog-cloud continuum, allowing compute, storage, and networking \textcolor{black}{capabilities from the fog to the objects} \cite{galambos2020cloud}. 

Another computing paradigm to emerge \textcolor{black}{from} edge computing is \textit{Mobile Edge Computing (MEC)} standardized by European Telecommunications Standards Institute (ETSI) \cite{hu2015mobile}. As of 2017, the ETSI renamed \textcolor{black}{MEC} to "Multi-Access Edge Computing" to \textcolor{black}{reflect growing} interest by non-cellular operators \cite{giust2018mec}. MEC is an edge computing technology that brings processing and storage capability to the network's edge \textcolor{black}{within} the Radio Access Network (RAN) to minimize latency and increase context awareness. MEC servers function in tandem with radio network controllers or macro base stations. The servers execute many instances of the MEC host, which may compute and store data across a virtualized interface. MEC reduces the computational and energy consumption of mobile devices by bringing the resources closer to the base stations, thus \textcolor{black}{enabling} one-hop services, minimizing the latency associated with data transmission.

Although originally developed to cater to the computational incapacity of cellular mobile devices like smartphones, \textcolor{black}{MEC} has been extended to the CPS systems. Since MEC \textcolor{black}{falls} under the \textcolor{black}{broader} umbrella of edge computing, we consider robotic applications using MEC \textcolor{black}{to be part of the Edge Robotics paradigm in our survey of state-of-the-art approaches}. MEC \textcolor{black}{provides a logical solution to enhance robot collaboration because it can not only offer efficient communication among robots but also efficiently distribute tasks}. The authors in \cite{chen2004studying} proposed a computational resource allocation strategy between mobile users and MEC servers to reduce task implementation delay and energy usage. Many such works utilizing MEC \textcolor{black}{are discussed} in Section \ref{sec:stateoftheart}. 

Further extending the particularities of MEC, \textit{Cloudlets} as envisioned by the researchers \cite{wang2020deep}, \textcolor{black}{are small clusters} or data centers capable of processing and storage that are located closer to mobile devices.\textcolor{black}{Often} referred to as mini-clouds, \textcolor{black}{cloudlets address the drawbacks associated with traditional cloud computing platforms}. The primary purpose of a cloudlet is to improve the interactive performance of mobile apps, particularly those with high end-to-end latency and jitter requirements. To meet the need for minimal delay, the proximity of cloudlets allows servers to deliver \textcolor{black}{highly} responsive cloud services to mobile nodes. The authors of \cite{afrin2018energy} target the energy consumption and service delay reduction of multirobot systems by distributing resources in a cloudlet-based emergency management service within a smart factory. 

Another concept to \textcolor{black}{emerge from} the umbrella of remote computational platforms is Dew computing, which is \textcolor{black}{beyond} the scope of our survey. 

\rowcolors{2}{gray!10}{white}
\begin{table*}[ht]
  \centering
  \renewcommand{\arraystretch}{1.3}
  \caption{Comparison Between Fog and Edge Robotics}
  \label{tab:fog_edge_robotics}
  \begin{tabular}{|p{3.2cm}|p{6.4cm}|p{6.4cm}|}
    \hline
    \rowcolor{gray!30}
    \textbf{Aspect} & \textbf{Fog Robotics} & \textbf{Edge Robotics} \\
    \hline
    Scope & Broader, covers intermediate layers between cloud and robots (e.g., gateways, local servers) & Narrower, focused on computing at or near the robot or sensor \\
    Location & Distributed across network nodes between cloud and robots & Located directly at or on the robots or edge devices \\
    Control \& Coordination & Managed across multiple network nodes, e.g., warehouses or local hubs & Mostly local control near the robot or sensor \\
    Latency \& Bandwidth & Reduces latency and bandwidth usage by intermediate processing & Minimizes latency by ultra-local processing near the robot \\
    Example Use Case & Warehouse robots coordinating via local servers with occasional cloud access \cite{song2025networked} & Robots processing sensor data on nearby embedded devices for real-time tasks \cite{zuo2025industrialinternetrobotcollaboration}\ \\
    Administrative Domain & Can span multiple trusted domains with policies on data flow & Typically operates within a single domain, such as a smart home or manufacturer \\
    \hline
  \end{tabular}
\end{table*}

\subsection{Difference between Cloud, Edge, Fog, MEC, Cloudlets}
The primary distinction between the aforementioned computing paradigms is that although \textcolor{black}{Cloud Computing} provides centralized processing, other computing paradigms, such as Edge, Fog, Mist, and Cloudlets, are built for distributed processing. Cloud computing allows for \textcolor{black}{high} scalability and \textcolor{black}{cost-effective} storage, but edge and other computing \textcolor{black}{architectures} may necessitate smaller, specialized processing equipment. Edge computing provides lower latency and response times, reduced power consumption and bandwidth costs, and enhanced security compared to the cloud. 

The terms ``Edge" and ``Fog" are \textcolor{black}{sometimes} used interchangeably due to conceptual similarities; however, there are important distinctions. Both permit data transmission from robots to the cloud, but edge computing allows complete, localized data processing at the edge, removing the need for a cloud entirely. In contrast, fog computing \textcolor{black}{acts} as a mediator between the edge and the cloud for data pre-processing \textcolor{black}{before cloud upload}. Edge robotics may entail connecting to the robot's sensors and controllers to deliver large amounts of data to the cloud, or \textcolor{black}{alternatively, this data may be processed entirely} at the network edge. The fog, on the other hand, is a computational layer that may accept data from the edge layer and analyze it before sending it to the cloud. Fog computing eliminates unnecessary data to reduce clutter in the cloud, offering \textcolor{black}{lower} latency and \textcolor{black}{better} efficiency of data traffic. However, one downside of fog computing is that it \textcolor{black}{cannot fully} replace edge computing, and the latter can be implemented without \textcolor{black}{the additional infrastructure overhead that fog requires}. Edge computing, on the other hand, has fewer peripheral layers than Fog and thus offers less scalability. It cannot provide resource pooling and interoperability \textcolor{black}{to the extent that fog can}. \textcolor{black}{Nonetheless, it relies} on multiple links for data transport, which is a significant disadvantage of fog computing. \textcolor{black}{In short, both edge and fog computing in robotics facilitate low-latency, secure, and efficient processing close to robotic platforms, enabling real-time tasks such as object detection and control. Edge computing primarily involves static infrastructure for computation offloading \cite{taleb2017multi}, whereas fog computing builds upon this by incorporating mobile and resource-constrained devices, offering enhanced flexibility, security, and context-awareness across the cloud-to-edge spectrum \cite{8304857}. Table \ref{tab:fog_edge_robotics} provides a side-by-side comparison of edge and fog computing in the context of robotics. For a deeper exploration of the challenges involved in integrating fog computing into robotic systems, we refer readers to \cite{9565454} and \cite{gudi2019fogroboticssummarychallenges}.}

For fog, MEC, and cloudlet paradigms, end devices and edge servers all install operating systems and \textcolor{black}{application-specific} software. Edge servers or the cloud serve as platforms for end devices to offload computation-intensive tasks to ensure timely data processing and energy efficiency. However, for fog and cloudlets, complete offloading is not feasible, as they are generally capable of executing only the necessary code rather than the entire program. Cloudlets mainly use virtual machines (VMs) for virtualization, while MEC and fog can also leverage containers for lighter-weight and faster deployment \cite{ren2019survey}. 

\section{Key Characteristics of Edge Robotics}
\label{sec:edge_robotics}
Edge Robotics possesses several characteristics that \textcolor{black}{mirror those of} edge computing. The following are the defining features of edge computing that make it beneficial \textcolor{black}{for deployment in} robotics applications.

\subsection{Proximity}
Edge computing makes remote computational resources and services available to the robots in close proximity, which can improve their performance for time-sensitive applications. The availability of computing resources and services \textcolor{black}{directly} at the edge of the network enables robots to use \textcolor{black}{context-aware network} information to make offloading and service-based decisions. Robots can be made more intelligent by utilizing the \textcolor{black}{remotely available services} to realize the concept of ``remote brain" for the efficient execution of logical algorithms. 

\subsection{Low latency}
Edge computing brings processing resources and services closer to robots, reducing latency in service access. Edge computing's low latency enables robots to run resource-hungry and delay-sensitive applications on resource-rich Edge devices (e.g., router, access point, base station, or dedicated server) for faster execution of algorithms like SLAM, navigation, object detection, gesture recognition, etc. 

\subsection{Geographical distribution}
Unlike cloud services, which are located far from local robots, edge computing brings the \textcolor{black}{compute infrastructure} closer by allowing similar service provisioning as the cloud. Their dense geographical distribution makes them an excellent contender for computational offloading in mobile robots, as it allows better fault tolerance or \textcolor{black}{resilience to} network loss through faster backups and recovery. 

\subsection{Mobility support}
Edge computing offers mobility by utilizing protocols like the Locator/ ID Separation Protocol (LISP) to interact directly with mobile devices or robots.  The LISP protocol separates \textcolor{black}{host identity from location identity} and creates a distributed directory system. The basic idea that enables mobility support in edge computing is \textcolor{black}{this decoupling, allowing mobile robots to access services closer to their current location, thereby enhancing performance}. 

\subsection{Heterogeneity}
The existence of various platforms, architectures, infrastructures, processing, and communication technologies employed by edge computing elements is referred to as heterogeneity in edge computing (end devices, edge servers, and networks). Edge devices can also be customized, adding further layers of heterogeneity, including software, hardware, and technological variances, APIs, custom-built policies, and platforms. In addition, \textcolor{black}{network-level heterogeneity may emerge}, leading to a variety of communication technologies that influence the delivery of edge services to the mobile robots.

\section{State of the Art: Edge Robotics} 
\label{sec:stateoftheart}
This section critically analyzes the literature on Edge Robotics paradigms, including Mobile Edge Computing (MEC) in robotics, with a focus on their objectives and outcomes.

\subsection{Computational Offloading}
By employing a multi-tier cloud and edge computing architecture to dynamically offload computationally expensive parts of the algorithm, Dey et al. \cite{dey2016robotic} \textcolor{black}{demonstrates} the usefulness of an edge computing platform for SLAM performance optimization. The edge server decides on offloading depending on compute and communication load, as well as energy utilization, with the goal of reducing execution time. To solve the SLAM challenge, the authors used a modified particle filter technique in conjunction with a dynamic offloading decision-making mechanism, evaluating the strategy through experiments conducted on VMs. The \textcolor{black}{results} revealed that the proposed dynamic offloading strategy \textcolor{black}{consistently} outperformed the static offloading, \textcolor{black}{significantly reducing} the mean execution time. However, no real-world testing was performed, where the communication layer \textcolor{black}{could} become a bottleneck due to high data transmission between robots and edge gateways.  

Dechouniotis et al. \cite{dechouniotis2022edge} \textcolor{black}{presented} a uniquely developed architectural testbed, NETMODE, based on the edge computing domain, \textcolor{black}{designed} to manage heterogeneous computing, network resources, and mobile robots \textcolor{black}{for complex} smart manufacturing scenarios. An edge-assisted SLAM application is deployed and tested using the suggested architecture \textcolor{black}{which advanced previous efforts through experiments with SLAM algorithm variations based on map update interval value (in seconds) implemented on local and edge devices provided a comparison of CPU use of the \textit{gmapping} method on the two resources, thus addressing earlier lack of real-environment testing}. The experimental findings showed considerable acceleration in the performance of compute-intensive workloads when edge infrastructure was used. 

Li et al. \cite{li2020intelligent} suggest an intelligent control approach \textcolor{black}{to meet the} computing power needs of AI high precision algorithms, particularly when a large number of robots create vast data in real time. The authors evaluated and compared existing working techniques for \textcolor{black}{robotic inspection of} power grid stations, and \textcolor{black}{introduced} an intelligent control strategy for live robot operation based on cloud and edge computing infrastructure. They analyzed the results of experiments on live working robots, which show that by using edge computing to filter data for the cloud platform, their proposed control method can meet the computational power requirements of AI deep learning, neural network coordinated control, self-learning control models, and so on, and can be extended to multiple large numbers of robots working simultaneously generating massive data. Their edge module \textcolor{black}{comprised} of several heterogeneous \textcolor{black}{components equipped with} GPU, NPU, TPU, APU, and FPGA capabilities. It processed sensor inputs, generated operation instructions, filtered status information and algorithm parameters, and received updates and training from the cloud. \textcolor{black}{Thus, their proposed work moved beyond SLAM to support scalable, real-time AI computation across robot fleets, enabled by a diverse hardware edge ecosystem.}


\textcolor{black}{A new perspective of integrating UAVs for adaptive coverage is proposed by} Wang et al. \cite{wang2021collaborative} thus shifting focus to a collaborative mobility-aware strategy. The proposed collaborative task offloading strategy uses improved genetic algorithms in mobile edge computing (MEC), introducing the unmanned aerial vehicle (UAV) cluster, addressing the limitations of fixed base stations in geographically constrained areas.
They put forward a collaborative task offloading model to offload tasks to UAVs or a base station. An objective function is put forward to jointly minimize task latency and energy consumption, and the genetic algorithm solves the resource optimization problem by proposing an optimal collaborative task offloading strategy. \textcolor{black}{Although simulation results provide supportive evidence for performance gains, the model's applicability under dynamic and unpredictable conditions in real-world scenarios remains unanswered.}

Recent work by Tahir and Parasuraman \cite{10473568} introduced a utility-aware dynamic task offloading strategy for multi-edge robotic systems, aiming to minimize task latency and enhance resource utilization for sequential multi-robot tasks. In another study \cite{10473544}, the authors proposed a decentralized, edge-enabled scheduling strategy for collaborative multi-robot systems, optimizing resource use and task performance, and demonstrating improvements in latency, throughput, and frame rates compared to traditional edge-based methods. \textcolor{black}{This work provides a distinctive perspective on multi-robot coordination and utility-driven decision-making under edge environments, moving from individual offloading to team-based offloading decisions.}

\textcolor{black}{Baruffa et al.\cite{10527382} add a communication-centric approach to the existing literature by proposing a testbed architecture that combines cloud/edge computing and multi-RAT (Radio Access Technology) networks for mobile robotic applications.
Their proposed work enables computational offloading for AI-driven robotic navigation via Kubernetes, Istio, and 5G/mmWave networks. The system is validated through a use case in which a ground robot offloads intensive vision tasks to distributed data centers in real time. Their work broadens the design space to include network heterogeneity alongside computing resources.}

\subsection{Context Awareness}
Klaas et al. \cite{Klaas2020} propose context-aware semantic path optimization for mobile robots that takes advantage of edge computing capabilities inside a distributed microservice-based autonomous control architecture. The objective is to add a semantic layer with a ROS-based modular navigation stack to include significant environmental data from RGB cameras into local path planning. Convoluted Neural Networks are used to semantically classify RGB pictures (CNNs). Similarly, to improve the quality of the computed path from ordinary local planners, several approach integrates robot policies based on the semantic classification of the robot's surroundings \cite{haseeb2018wisture,kannan2020material}. These works only emphasize their importance by exploiting edge computing for the goal of embedding semantics into navigation stacks, but no significant improvement in decreasing operational time was assessed.

Likewise, the same authors, Lambrecht et al. \cite{Lambrecht2019EdgeEnabledAN} presented a micro-service-based architecture for autonomous mobile robots that offloads the entire navigation stack towards the edge and found advantage of edge computing in terms of energy consumption and on-board computational requirements, however, offloading of navigation took a longer operation time for them. 

The authors Antevski et al. \cite{Antevski} present an advanced control algorithm in MEC to control the robot speed remotely. They begin by running a series of tests to determine the relationship between robot control latency and Wi-Fi signal intensity. The resultant characterization was utilized as a starting point for creating, implementing, and testing a control algorithm that uses context information from the Wi-Fi signal to adjust the robot's speed for smoother driving. The results of their experiments indicate that adjusting the robot's speed depending on the Wi-Fi signal given by the MEC information service may successfully generate smoother driving at high speeds. This enhancement enables the robot to perform faster than if no context information is consumed. \textcolor{black}{The proposed approach advanced the real-time adaptation of edge-robot control systems from compute-centric approach to a communication context-aware autonomy.}
 
A context-aware testbed architecture, COTORRA, proposed by Groshev et al. \cite{Groshev2021} operates on the edge of the network. In addition to offering deployment of serverless plugins, it supports testing of time-sensitive applications. It can also emulate unpredictable network conditions that enable a realistic environment where plugins can be tested. They validated through experiments the feasibility of their architecture by running an orchestration algorithm that interacts with the network infrastructure and mobile robots, and performed autonomous navigation that improved task latency to less than 15ms.  \textcolor{black}{The authors claim COTORRA to be the first edge/fog-based robotic testbed of its kind. The approach solidifies testing of context-aware edge-based robotic control under variable network conditions compared to specific controlled conditions employed by previous works.}

\textcolor{black}{In the 6G communication domain, Zeng et al. \cite{10804598} present a semantic communication (SemCom) framework tailored to robotic edge-AI systems where a robot collaborates with a remote edge server acting as its “brain.” The system uses a Knowledge Graph (KG) to represent task-relevant object-action sequences, known as Knowledge Paths (KPs), enabling the server to guide the robot in achieving specific goals based on its environmental observations. A key innovation is a Robotic SemCom protocol that includes semantic matching between KG elements and server-side classifiers, enabling the robot to stream compressed feature data to the server for real-time object recognition and path identification. This work primarily meets ultra-low-latency requirements through a feature transmission scheme (ULL-FT) that leverages the classifier's robustness to tolerate channel errors, achieving a communication-efficient balance between reliability and speed.  The effectiveness of the proposed approach is demonstrated using synthetic and real-world datasets, showing significant latency reductions without compromising inference accuracy. Overall, the work pioneers a goal-oriented air interface for 6G robotic applications where task-driven communication is prioritized over raw data fidelity.}

\subsection{Localization}

According to the authors of \cite{ThongTran2020}, it is critical to use edge computing and cloud services to create a well-developed Autonomous Mobile Robot (AMR) system since edge and cloud may minimize AMR power consumption and make integration with sensors or \textcolor{black}{IoT} devices easier. They create\textcolor{black}{d} an efficient AMR system from scratch, using an NVIDIA Jetson TX2 as an edge computing power-efficient module that can handle the operation of AMR in unstable or unreliable network conditions, a serverless architecture based on Amazon Web Services (AWS) for Cloud computing, and a user interface built in React Native They evaluated the performance of their systems by running an indoor delivery service, a face-recognition function for identifying sender and recipient, and caching all requests and responses via cloud-based microservices for rapid task execution. 

Many seminal works in tackling the resource constraints posed by local onboard computing in executing SLAM algorithms have turned to edge computing to improve efficiency and reduce task latency. P. Huang et al. present a collaborative multirobot laser SLAM \cite{Huang2021} that leverages the edge computing concept for SLAM execution optimization. The robot-edge synergy uses edge computing to enable robots to execute SLAM and produce a global map. Previous measurements on total SLAM offloading on an Alibaba Cloud server show\textcolor{black}{ed} a significant transmission delay. Because feature extraction, matching, and transform optimization are critical submodules of map fusion, the core of multirobot SLAM, they consume more than 70\% of \textcolor{black}{the} processing time.  ColaSLAM addresses this issue by deploying map fusion and mapping modules to edge servers to build a number of sub-maps and conduct feature extraction and matching by recognizing overlaps between these maps, resulting in map fusion. They propose an adaptive map redistribution approach to speed up map fusion among several edge servers. 

A coordinator generates an undirected graph composed of submaps while taking into account the overlapping degree and bandwidth condition, then splits the graph and generates an offloading decision in which each edge cluster is assigned a submap for execution, and finally, the edge server merges the received sub-maps and uploads the results to the cloud center to minimize the latency from data collection to the cloud side. Experiments on simulation and prototype evaluate their results on workload balancing against greedy and random approaches and latency, and realize that the proposed algorithm can reduce the processing latency up to 40\% compared to the direct offloading to the cloud and brings up to 52\% on average improvement compared to random and up to 15\% against the greedy algorithm.  

\textcolor{black}{
Same authors propose RecSLAM \cite{9693970}, a hierarchical robot–edge–cloud system designed to enable low-latency, multi-robot laser SLAM. It addresses the limitations of traditional SLAM approaches—such as excessive local computation and high cloud offloading latency—by distributing SLAM tasks across nearby edge servers. Through real-world experiments and simulations, the system demonstrates significant improvements in efficiency and latency reduction.}

\textcolor{black}{Liu et al. \cite{10472967} propose a centralized multi-robot SLAM system using a robot-edge-cloud architecture, where edge servers handle lightweight optical flow tracking between non-keyframes by receiving only keypoints, reducing computation on robots. Keyframes are compressed and sent to the cloud for pose estimation and map fusion. This approach minimizes communication load while maintaining SLAM accuracy.}

Researchers Sarkar et al. in \cite{Sarker2019} solve the SLAM optimization issue with energy efficiency and performance guarantees as key objectives by leveraging both the Edge and Fog layers for computationally heavy tasks while leaving the Cloud primarily for monitoring, control, and visualization. They suggest a system architecture with four layers: a robot layer and an edge layer for data processing, a fog layer for distributed storage, and a cloud layer for monitoring and general mission control. When the robot moves from one edge to another, the Edge layer takes a real-time decision to reduce delay by sharing the computing load of the robot, while the Fog layer conducts a handover mechanism with negligible data loss.

This approach includes exchanging prior map data created by the robot during SLAM edge handover. Robots save energy by sending data to the gateway, i.e., the Edge and Fog layer, rather than directly to the Cloud. They observe decreased execution delay due to a shorter data path across the entire cycle, improvement in network bandwidth as possible bottlenecks are addressed by avoiding robot-to-Cloud data transmission, and considerable energy improvement on the robots through trials. Although they conducted real-world testing on a small automobile prototype, the results were insufficient to demonstrate the efficacy of the suggested method.
Perhaps some stress testing on a real-world multi-robot system may aid in further evaluating the algorithm.

A 3D semantic map creation approach for mobile robots based on enhanced ORB-SLAM2 is proposed by Cui et al. \cite{Cui2020} to ensure that mobile robots have a certain capacity to perceive targets in indoor environments and to improve the degree of intelligence of robots. The proposed technique involves first the implementation of an improved version of YOLOv3 to detect indoor objects, then a real-time semantic segmentation network model is employed based on deep learning that segments the objects to classify their pixel points on a two-dimensional image and to improve the accuracy of semantic segmentation, the BAFF feature fusion algorithm is applied. A 3D semantic map is then created by the use of SLAM that estimates the pose of the image as a result of semantic segmentation and using this depth information, projects it onto the 3D environment.

Solving the resource optimization problem on robots, \textcolor{black}{the} authors Ben Ali et al. \cite{10.1145/3561972} propose the implementation of ORB-SLAM2 using edge computing to offload parts of Visual-SLAM. They split the architecture of the algorithm between the edge and the robot by keeping the tracking module on the robot while moving more intensive computational modules, i.e., local mapping and loop closure, to the edge. By keeping the components tightly connected, they hope to achieve their goal of lowering computational and memory overhead without compromising performance precision.

Because the local mapping and loop closing modules rely on the global map for some of their computation, it is critical to update the local map whenever the global map changes, necessitating close communication between the modules, which is accomplished by introducing three separate network connections. The first two connections are responsible for getting data from the tracking module; i.e., one where the tracking module communicates the features and local geometry in \textit{Frames}, the other is to pass the keyframe if tracking decides to create a new keyframe, which \textcolor{black}{is} used for creating and storing the global map on the edge. The third connection is for updating the local map on the mobile robot and a map synchronization module on the edge that synchronizes the local map updates with the latest optimized changes.

Upon implementation on a prototype through rigorous experiments, the split architecture proposed reduces computation and memory cost while keeping the execution time to a minimum. The latency involved with an update of the local map between the robot and the edge also substantially reduces the overall VSLAM execution time. They estimate their performance through mapping accuracy by mean localization error, and observed negligible loss of accuracy in the final map and the trajectory taken.

Another work by Chen et al. \cite{Chen2021} investigates edge-computed aided autonomous flight for UAV (ECAAF), in which vision and communication modules deployed in the framework interact and support each other using edge computing and offloading to speed up the UAV mission. ECAAF has three functionalities: 3D map acquisition with an edge layer, radio map generation from the 3D map, and live trajectory planning. A positive feedback loop also allows interaction between edges and UAVs for video offloading, 3D map quality, and channel state of the trajectory form. They test their suggested strategy using simulations, which show that it can increase mission performance by boosting interconnectivity.

\rowcolors{2}{gray!10}{white}
\begin{table*}[ht]
  \centering
  \renewcommand{\arraystretch}{1.3}
  \caption{Comparison of Edge Robotics applications based on objectives}
  \label{tab2}
  \begin{tabular}{@{}p{3.2cm}lllllll@{}}
    \hline\hline
\rowcolor{lightgray}
\textbf{Reference} & 
\parbox[c]{1.5cm}{\centering Computational\\Offloading} & 
\parbox[c]{1.5cm}{\centering Context\\Awareness} & 
\parbox[c]{1.5cm}{\centering Localization} & 
\parbox[c]{1.5cm}{\centering Navigation} & 
\parbox[c]{1.5cm}{\centering Minimizing\\Latency} & 
\parbox[c]{1.5cm}{\centering Resource\\Optimization} & 
\parbox[c]{1.5cm}{\centering Minimizing\\Energy\\Consumption} \\
    \hline
    S. Dey et al.\cite{dey2016robotic} & \cmark & \xmark & \cmark & \xmark & \cmark & \xmark & \cmark\\ \hline
    D. Dechouniotis et al.\cite{dechouniotis2022edge} & \cmark & \xmark & \cmark & \cmark & \cmark & \cmark & \xmark \\ \hline
    W. Li et al. \cite{li2020intelligent} & \cmark & \xmark &\xmark &\xmark &\xmark &\xmark  &\cmark \\ \hline
    P. Wang et al. \cite{wang2022design} & \cmark & \xmark &\xmark &\xmark &\xmark&\xmark &\xmark \\ \hline
    H. Wang \cite{wang2021collaborative} & \cmark & \xmark & \xmark & \xmark & \cmark & \cmark & \cmark \\ \hline
    N. Tahir et al.\cite{10473568, 10473544}& \cmark & \xmark & \cmark & \cmark & \cmark & \cmark & \xmark \\ \hline
    \textcolor{black}{Baruffa et al. }\cite{10527382} & \cmark & \xmark & \xmark & \cmark & \cmark & \cmark & \xmark \\ \hline
    T. Klaas et al.\cite{Klaas2020} & \cmark & \cmark & \xmark & \cmark & \xmark & \xmark & \xmark \\ \hline
    J. Lambrecht et al.\cite{Lambrecht2019EdgeEnabledAN} & \cmark & \cmark & \xmark & \cmark & \xmark & \xmark  & \cmark \\ \hline
    K. Antevski et al.\cite{Antevski} & \xmark & \cmark & \xmark & \cmark &\cmark &\xmark &\xmark \\ \hline
    M. Groshev et al. \cite{Groshev2021} & \cmark & \cmark & \xmark & \xmark & \cmark & \xmark & \xmark \\ \hline
    \textcolor{black}{Q. Zeng et al.}\cite{10804598} & \cmark & \cmark & \xmark & \xmark & \cmark & \cmark & \xmark \\ \hline
    T. Thong Tran et al.\cite{ThongTran2020} & \cmark & \xmark & \cmark & \cmark & \cmark & \xmark & \cmark \\ \hline
    P. Huang et al. \cite{Huang2021,9693970} & \cmark & \xmark & \cmark & \xmark & \cmark & \cmark & \xmark \\ \hline
    \textcolor{black}{Lui et al.} \cite{10472967} & \cmark & \xmark & \cmark & \xmark & \cmark & \cmark & \xmark \\ \hline
    V. Sarkar et al.\cite{Sarker2019} & \cmark & \xmark &\cmark & \xmark & \cmark & \cmark & \cmark \\ \hline
    X. Cui et al.\cite{Cui2020} & \xmark & \xmark & \cmark & \xmark & \xmark & \xmark & \xmark \\ \hline
    A. Ben Ali et al.\cite{10.1145/3561972} & \xmark & \xmark &\cmark &\cmark &\cmark & \xmark &\xmark \\ \hline
    Q. Chen et al.\cite{Chen2021} & \xmark & \xmark & \cmark & \cmark & \cmark & \xmark & \xmark \\ \hline
    U. Palani et al.\cite{Palani2022} & \xmark & \xmark & \xmark & \cmark & \cmark & \xmark & \xmark \\ \hline
    Z. Fan et al.\cite{Fan2019} & \xmark & \xmark & \cmark & \cmark & \cmark & \xmark & \xmark \\ \hline
    J. Li et al.\cite{Li2021} &  \xmark & \xmark & \xmark & \cmark & \xmark & \xmark & \xmark \\ \hline
    L. Qingqing et al.\cite{Qingqing2019} & \cmark & \xmark & \cmark & \cmark & \cmark & \xmark & \xmark \\ \hline
    S. Hayat et al.\cite{Hayat2021} & \cmark & \xmark & \cmark & \cmark & \cmark & \xmark & \cmark \\ \hline
    \textcolor{black}{G. Li et al.} \cite{li2024edge} & \cmark & \xmark & \xmark & \cmark &\cmark & \cmark &\xmark \\ \hline
    C. Asavasirikulkij et al.\cite{Asavasirikulkij2021} & \xmark & \xmark & \cmark & \cmark & \cmark & \xmark & \cmark \\ \hline
    R. Yin et al.\cite{Yin2022} & \cmark & \xmark & \xmark & \xmark & \cmark & \cmark & \cmark \\ \hline
    \textcolor{black}{N. Tahir et al. }\cite{10757784} & \cmark & \xmark & \xmark & \cmark &\cmark & \cmark & \xmark \\ \hline
    \textcolor{black}{K. Chen et al.} \cite{10610759} & \cmark & \xmark & \xmark &\xmark & \cmark & \cmark & \xmark \\ \hline
    L. Qingqing et al.\cite{Qingqing} & \cmark & \xmark & \xmark & \xmark & \cmark & \cmark & \xmark \\ \hline
    D. Spatharakis et al.\cite{Spatharakis2022} & \cmark & \xmark & \cmark & \cmark & \cmark & \cmark & \xmark \\ \hline
    X. Huang et al. \cite{Huang2021x} & \cmark & \cmark & \xmark & \xmark & \cmark & \xmark & \xmark \\ \hline
    \textcolor{black}{S. Bouhoula et al.}\cite{10811394} &  \cmark & \xmark & \xmark &\xmark & \cmark &\cmark & \cmark \\ \hline 
    Wang et al. \cite{9369456} & \cmark & \xmark & \xmark & \xmark & \cmark & \xmark & \cmark \\ \hline
    F. Farahbaksh et al. \cite{Farah} & \cmark & \cmark & \xmark & \xmark & \cmark & \xmark & \cmark \\ \hline
    \textcolor{black}{Zeng et al.} \cite{10558816} & \cmark & \xmark & \xmark & \xmark &\xmark & \cmark & \cmark \\
    \hline\hline
  \end{tabular}
  \caption*{\footnotesize{\textcolor{black}{\textbf{Legend:} \cmark~Yes indicates that the objective is fully or partially explored in the research; \xmark~No indicates that the objective is not considered.}}}
\end{table*}

\subsection{Navigation}
Many works in the literature attempt to address the MRS navigation issue using edge computing. Palani et al. \cite{Palani2022} propose an edge computing-based autonomous robot, where the edge is deployed for data processing to calculate the path to the source of product boxes in a manufacturing industry scenario to avoid accidents involving product damage. They develop a six-degree-of-freedom manipulator in SolidWorks and use it to test a navigation algorithm in an obstacle-filled environment.

The edge computing module used in their suggested system intends to solve the process of discovering the path for navigation utilizing a small quantity of data at a fast speed and with minimal time delay by employing obstacle detection and avoidance, as well as enabling direct control. Experiments are carried out to verify the accuracy of direction tracking and the time of operation of the robot, but no important conclusions can be derived because the research lacks comparison with benchmark methodologies, \textcolor{black}{limiting the study's impact}. 

Fan et al. \cite{Fan2019} offer a collaboration method based on an edge computing framework, a robot "StellaX," and a manipulator in an attempt to accomplish navigation and object identification for grasping through cooperation across heterogeneous robots. The mobile robot "StellaX" features three omnidirectional wheels and is outfitted with a LIDAR scanner for SLAM and navigation, as well as a stereo camera for object identification and grasping. The manipulator and the robot communicate via edge servers, where information about the object used for grasping is saved and transferred so that the manipulator may access it. 
Docker Edge Robotics Framework (DERF) is a container engine technology based on Linux container (LXC) that provides the benefits of quick and efficient deployment, high resource utilization, and simple administration. DERF is \textcolor{black}{divided} into the following categories: service provider, service requester, cloud center, edge node, and physical layer.

A Docker container image is built and registered with the cloud center in order to carry out the navigation (\textit{gmapping} with \textit{AMCL} ROS package as described in the article).  As the requester, the user can send the service request over the network, and once received, the management module in the cloud center performs a service query in the image repository to confirm that the service exists and the robot is idle, after which the image is instantiated to form a functional application that will send task instructions to the physical layer via edge node. Experiments on object identification and navigation demonstrate the differences between local, cloud, and edge computing. The entire execution time was greatly decreased by using edge computing, and the upload time for edge computing is always less than that of the cloud. The use of the edge layer also considerably decreased CPU and memory \textcolor{black}{usage} rates, as well as navigation execution time.

Authors Li et al. \cite{Li2021} propose a visual navigation algorithm implemented on the edge computing platform for agricultural robots based on deep learning image understanding. To process pictures acquired by the proposed vision system, the approach initially employs a cascaded deep convolutional network and a hybrid dilated convolution fusion algorithm. The modified Hough transform technique is then used to extract the \textcolor{black}{path from} processed pictures. At the same time, the agricultural robot's posture is being adjusted to enable autonomous navigation. Their proposed method is validated using non-interference and noisy experimental scenes in real-world experiments. The results of the experiments reveal that the algorithms proposed performed better in the non-interference scene and complex noise scenes, but no experimental data can be obtained from using an edge computing platform or any knowledge of whether their objectives were satisfied through the use of edge devices. 

Qingqin et al. \cite{Qingqing2019} investigate the state-of-the-art in visual odometry-based autonomous navigation for MRS that are confined by computing and sensor resources.
Image compression reduces the quality of navigation, but it is not the \textcolor{black}{only method for improving execution time and accuracy}. Through this experimental study, they discovered that image size and network bandwidth may be reduced by an order of magnitude without compromising the accuracy of the odometry methods, even in challenging environments. Their tests look at two parameters: network latency and odometry algorithm accuracy. Using the EuRoC dataset, they examine the processing time required for the feature extraction and posture estimation processes for each of the \textcolor{black}{image} compression ratios. 

Their findings show that reducing picture quality to a certain amount has no effect on \textcolor{black}{odometry} accuracy, indicating that the system can benefit from a computational offloading approach using edge computing. They demonstrate through findings that offloading does not cause odometry delays and actually enhances frame rate performance when powerful edge gateways are deployed. They suggest an edge computing offloading strategy that can provide several benefits to a large MRS system, including reduced cost and energy consumption, as well as improved performance and reliability. 

Hayat et al. \cite{Hayat2021} analyze the importance of edge computing in 5G for vision-based navigation algorithms for drones. They deployed edge devices to offload computing tasks for image processing. They test various modes of computational offloading, including onboard, fully offloaded to the edge, and partially offloaded. Their findings reveal that for all image resolutions, the image processing time for complete offloading is longer than for partial offloading, and this difference gets more pronounced as resolution increases. When the computational power of the server is twice that of the drone, partial offloading results in faster image processing times than onboard processing. This is not true for complete offloading, however. Full offloading necessitates long image transmission times, and increased edge compute power has little effect on overall image processing time.

The authors then investigate the effect of 5G data rates on image processing time in each mode. They conclude that higher up-link speeds can meet the image transmission requirements of the full offloading mode.  Extracting features onboard and sending them to the edge provides more advantage than delivering the entire image. They also highlight that partial offloading will place less strain on the communication network in terms of transmission rate than full offloading, but still needs some onboard processing. 

\textcolor{black}{Li et al. \cite{li2024edge} propose an Edge Accelerated Robot Navigation (EARN) framework that enhances robot navigation by adaptively switching between local and edge-based motion planning based on resource availability. Unlike traditional approaches that focus solely on algorithm design, EARN integrates decision-making and motion planning under communication and computation constraints using model predictive switching. It effectively utilizes edge resources to enable advanced collision avoidance and optimized path planning. While the framework demonstrates strong performance in both simulated and real-world settings, its reliance on low-latency edge connectivity and complex optimization algorithms may pose deployment challenges in dynamic or resource-constrained environments.} 

\subsection{Minimizing task latency}
Minimizing task latency is \textcolor{black}{largely} attributed to improving the network conditions in the edge computing domain.
Asavasirikulkij et al. \cite{Asavasirikulkij2021} devise a new peer-to-peer wireless communication platform for communication between a mobile robot and an edge device in a smart factory scenario.  The application tested includes a mobile robot equipped with sensors and connected to an edge device and a serial link manipulator. Through their proposed wireless platform that offers high bandwidth and low communication latency \textcolor{black}{via} the use of \textcolor{black}{a} Software-Defined Wireless Networking (SDWN) approach, they implement a pose estimation through SLAM \textcolor{black}{on} the edge device based on LIDAR scan data transmitted from the robot to the edge.

They extend their work further by proposing a wireless network connecting several machines in an automation cell, i.e., a robot, a manipulator, a warehouse, and an edge server. The edge is deployed to collect all the data from the machines in-house to generate an updated map for controlling the automation cell during robot collaboration. They test their proposed setup through a use case scenario where the mobile robot picks up a package from the manipulator and delivers it to the warehouse using the pose estimates provided by the edge device in the automation cell. Experiment\textcolor{black}{al} results evaluate the collaborative map generated as well as the path execution comparison between the planned path and the mobile robot's actual path used for navigation. Compared to two other benchmark techniques, their proposed architecture undergoes less processing latency to compute SLAM. 

Another work put forth by Yin et al. \cite{Yin2022} elaborates on a MEC-based multirobot cooperation (MRC) system with Master Robot (MR) and Slave Robots (SRs). They envision the MR as an edge device in charge of making decisions based on sensing conducted by the SRs. In addition to sensing data, the SRs offload various functions to the MR in order to reduce energy utilization \textcolor{black}{onboard}. The research provides two resource management solutions for minimizing and balancing energy usage across multiple SRs while ensuring task completion on time. The energy consumption of the SRs is minimized and balanced in the first scheme to increase the function duration of the MRC systems. 
The residual energy of the SRs is employed in the second scheme to deal with the unpredictable wireless communication environment. As a result, instead of transferring the data to the MR, each SR may handle it locally. The latter scheme is more durable than the first, although it may consume more energy. They test their methods using simulations, which show that MEC can help MRC handle time-critical heavy computing tasks effectively.  

The mobility of robots poses an additional challenge for computational offloading, particularly when multi-robot systems depend on real-time decision-making from AI systems deployed on edge servers. Frequent access point handovers lead to service migration between edge servers, resulting in increased latency and negatively impacting the performance of tasks such as autonomous navigation.

Tahir and Parasuraman \cite{10757784} propose a deep reinforcement learning framework based on the Deep Deterministic Policy Gradient (DDPG) algorithm to address the joint problem of task migration and access-point handover in vehicular networks. They introduce a joint allocation method for communication and computation across access points to minimize computational load, service latency, and interruptions, with the goal of maximizing Quality of Service (QoS). Their framework, evaluated through simulated experiments, enables smooth task switching among edge servers, reducing latency and improving service efficiency.

\textcolor{black}{The authors K. Chen et al. \cite{10610759} introduce the FogROS2-Latency-Sensitive (FogROS2-LS) framework, which dynamically selects the most suitable cloud or edge server to minimize latency for robots offloading computationally intensive tasks. Their framework offloads conventional on-board state estimators and feedback controllers to cloud and edge devices without requiring modifications to existing ROS2 applications. When multiple identical services are available, FogROS2-LS dynamically identifies and switches to the optimal service deployment that meets latency requirements, enabling robots with limited on-board computing power to safely and efficiently navigate dynamic, human-populated environments.}

Because the majority of the works in the literature involving computational offloading to edge computing focus on latency minimization, either in terms of execution time or transmission time, such studies are not repeated in this section for the sake of brevity.

\subsection{Resource Optimization}

Qingqing et al. \cite{Qingqing} suggest using FPGAs at the edge for computational offloading with low latency and high parallelism in a multirobot feature-based lidar odometry system. In three phases, their approach performs feature-based lidar odometry: The extraction of features from lidar data comes first. These characteristics might be geometric in form or stable points observed in two successive sweeps. Second, it employs feature correspondence by calculating the position difference between the sweeps.

Finally, the time interval between two successive sweeps is used to estimate the LIDAR movement. The purpose of employing FPGAs is to process scan data in real time with minimum computational resources and to parallelize data processing. They compare their technique to their own initial and unoptimized design, in terms of resource usage and concurrency gained, which was shown to be superior for the optimized proposed version.

Spatharakis et al. \cite{Spatharakis2022} provide a set-based estimation approach for the robot offloading mechanism in the context of edge robotics. Using a case study of a unicycle robot performing navigation and path planning, they assess the tradeoff between performance and resource consumption. An offloading strategy is \textcolor{black}{introduced} to compensate for the uncertainty of local estimation techniques with more accurate remote sensing, while balancing navigation accuracy and mission duration. \textcolor{black}{While traversing its environment, the unicycle robot uses remote estimates to correct inaccuracies in local estimates}. Offloading is performed via a framework that considers network conditions and computational availability while maintaining the stability of the control system.

\textcolor{black}{Their} decision-making computational offloading strategy is framed as a utility-based function that incorporates two localization techniques: an error-prone one, i.e., odometry-based localization (executed locally), and a more accurate, computationally intensive vision-based localization method executed on the edge server. They compared their utility-based offloading scheme with two others—exclusively local implementation and exclusively remote—in terms of average mission duration (in seconds), average offloading triggers, and success rate. They also presented a comparison of their utility-based offloading scheme under five different time-triggered settings, where remote estimation is triggered every $\mu$ seconds across a total of 35 experiments. Their technique outperformed all others by providing convergence guarantees while keeping mission duration low.

Huang et al. \cite{Huang2021x} \textcolor{black}{develop} an offloading service provider in Collaborative Vehicular Edge Computing (CVEC) system by launching an MEC server and scheduling PVs (Parking vehicles) on demand to perform offloading tasks.
To optimize network-wide task scheduling, efficient workload allocation and user-centric utility maximization are studied. Each task's offloading destination is determined probabilistically in a dynamic context. When necessary, the offloading service provider acts in the capacity of an offloading user to \textcolor{black}{initiate} a contract-based incentive mechanism for PVs. The authors base the offloading user's subjective assessments of utility in computational offloading using contract theory and prospect theory, and develop an ideal contract to maximize subjective utility under formation asymmetry. Numerical data collected through experiments indicate the proposed scheme's efficacy and efficiency.

\textcolor{black}{Bouhoula et al. \cite{10811394} present a DRL-based task scheduling strategy designed for MEC-enabled robotic systems in dynamic environments by jointly optimizing task execution, robot trajectory planning, and energy consumption, addressing the challenges caused by robot mobility, transmission delays, and resource constraints. The authors transform the problem into a Markov Decision Process (MDP), enabling real-time adaptive decisions such as local processing versus offloading, and device-to-device (D2D) task sharing. Their extensive evaluations demonstrate significant improvements over baseline methods in reducing delays and energy use. While the approach excels in scalability and adaptability, it remains dependent on simulation scenarios and may face challenges in deployment complexity in the real world.}

\subsection{Minimizing energy consumption}
Wang et al. \cite{9369456} \textcolor{black}{propose} a remote computational offloading mechanism for a swarm of robots connected to an edge server. The primary goal of their method is to reduce energy consumption and task execution time. They developed a deep reinforcement learning system for task scheduling that takes the robot's mobility into account. They showed the efficiency of their technique in decreasing energy consumption and delay through simulation. \textcolor{black}{However, their} system paradigm is based on a single edge server for management and compute offloading, \textcolor{black}{rendering the study incomplete} with respect to the deployment of multiple edge devices and their impact on the performance. 

Farahbaksh et al. \cite{Farah} present a computational offloading approach based on Bayesian learning automata for mobile edge computing. Their \textcolor{black}{method incorporates} contextual information awareness \textcolor{black}{which influences} decision-making for computational offloading from mobile devices to edge servers. Context information is captured via a control loop. Several metrics \textcolor{black}{are studied} to evaluate the proposed scheme, \textcolor{black}{including} energy consumption, execution time, resource utilization, network usage, task latency, module size, and offloading intervals. 

\textcolor{black}{Zeng et al. \cite{10558816} address the problem of aggregated and centralized training which strains on long-range backhaul transmission. They propose a collaborative edge training framework that leverages idle resources from trusted edge devices as a resource pool for efficient and sustainable training of large AI models. The authors design the system architecture, analyze scheduling strategies for energy optimization, and evaluate different parallelism methods using realistic testbeds.}

There is some overlap \textcolor{black}{among} the aforementioned efforts that address energy consumption using edge robots. Although not exhaustive, this list omits certain works for brevity. 

\section{Research Gaps and Limitations}
\textcolor{black}{Real-world evaluations of edge-assisted robot systems are critical for validating performance, yet many existing studies exhibit significant limitations in scope, generalizability, and experimental rigor. For example, Dey et al. \cite{dey2016robotic} conducted comprehensive hardware evaluations but relied solely on virtual machines as edge devices, with no physical robot deployment. Their experiments assumed idealized network conditions—constant bandwidth, uniform processing capabilities, and negligible variability—thereby overlooking crucial factors such as real-time network dynamics, energy constraints, and fault tolerance that are essential for practical SLAM offloading in mobile robotics.}

\textcolor{black}{Similarly, the NETMODE testbed by D. Dechouniotis et al. \cite{dechouniotis2022edge} employs Raspberry Pi 3 units as mobile robots, which offer limited computational capacity. While the system evaluates CPU usage, it omits other essential performance indicators such as latency, task completion success, localization accuracy, and energy consumption. Moreover, their manually configured architecture lacks support for dynamic task orchestration, scalability to multi-robot systems, or resource-aware decision-making at the edge, severely limiting its applicability to complex, long-term, or large-scale robotic deployments.}

\textcolor{black}{In another instance, the cloud-edge hybrid architecture proposed in \cite{li2020intelligent} reduces latency by offloading lightweight AI inference to the edge while relegating heavy model training to the cloud. While effective for improving processing speed, the system’s dependence on stable connectivity and the limited computational capacity of the edge restricts its robustness and scalability in dynamic environments, especially those involving multiple robots or unpredictable wireless conditions.}

\textcolor{black}{P. Wang et al. \cite{wang2022design} utilize a greedy task allocation strategy based on proximity (e.g., up/down/left/right neighbors), which, while computationally simple, does not scale to complex environments or guarantee globally optimal solutions. Their real-world experiments are confined to static and known layouts with fixed obstacles, lacking dynamic obstacle avoidance, uncertainty modeling, or real-time adaptive behavior. Similarly, H. Wang et al. \cite{wang2021collaborative} evaluate performance under static terminal assumptions, predictable task distributions, and idealized communications, which fail to reflect the nuanced variability of bandwidth, mobility, and hardware resource contention inherent in real-world mobile robotic systems.}

\textcolor{black}{Baruffa et al. \cite{10527382} present a modular testbed integrating 5G, mmWave, and cloud/edge infrastructure for robotic navigation, yet the system relies on specialized infrastructure and evaluates only a narrow slice of performance metrics. Notably, it lacks comprehensive testing for scalability, security, and adaptability to diverse deployment scenarios. Likewise, the work by J. Lambrecht et al. \cite{Lambrecht2019EdgeEnabledAN} shows impressive CPU and RAM offloading benefits and up to 38.59\% energy savings during navigation and vision tasks. However, these benefits are measured under controlled conditions—fixed robot velocity, no payload, and static workload—without accounting for edge-side energy costs or network-induced overheads. Moreover, the system lacks stress testing under fluctuating workloads, making it difficult to generalize to unpredictable operational contexts.}

\textcolor{black}{The COTORRA testbed \cite{Groshev2021} suffers from reliance on synthetic network variability via NetEm, using artificially injected delays rather than real-world wireless channel dynamics. Conducted in constrained indoor corridors, the experiments omit outdoor, user-interactive, or heterogeneous multi-robot environments. Orchestration evaluations are limited to a strict 15 ms latency target without assessing robustness against threshold violations, and the DLT federation component exhibits a prohibitively high 19-second delay—unsuitable for time-sensitive applications. Robot autonomy was limited to pre-programmed paths, lacking responsiveness to environmental complexity or uncertainty.}

\textcolor{black}{The ColaSLAM system \cite{Huang2021} successfully demonstrates collaborative multi-robot mapping using tiered edge-cloud fusion, significantly reducing on-robot computational load. However, it assumes uninterrupted wireless connectivity and performs no local data preprocessing, increasing transmission demands and vulnerability to network disruptions. Additionally, the system’s coordinator operates intermittently, which may introduce scheduling delays in dynamic conditions. Real-world tests also reveal mapping inaccuracies at boundaries and dependence on precise overlap detection, both of which can hinder map fusion reliability.}

\textcolor{black}{Finally, the Edge-SLAM framework by A. J. Ben et al. \cite{10.1145/3561972} is evaluated using a minimal hardware setup involving only two mobile devices and one edge server, which limits the generalizability of results to more complex settings. The use of replayed datasets over private Wi-Fi networks and a wired edge connection fails to capture performance under variable mobility, fluctuating bandwidth, or real-time wireless interference. While Edge-SLAM demonstrates accuracy comparable to ORB-SLAM2, architectural decomposition and latency could exacerbate drift in longer or more complex deployments. Moreover, tracking sensitivity issues inherited from ORB-SLAM2 remain unaddressed, highlighting gaps in robustness for real-world SLAM scenarios.}

\textcolor{black}{As evident from the critical examination of existing experimental frameworks, edge robotics has made substantial strides through innovative system architectures, distributed offloading mechanisms, and the integration of heterogeneous computing platforms. These developments have improved real-time processing for tasks such as SLAM and autonomous navigation, particularly in controlled environments. However, the evaluation of these systems often falls short of capturing the operational complexity of real-world deployments. Many experimental setups rely on idealized assumptions—such as stable, high-throughput wireless networks, deterministic robot behaviors, or fixed environmental layouts—that do not reflect the variability and uncertainty encountered in practical field conditions. Furthermore, reliance on small-scale testbeds, synthetic data replay, and limited performance metrics hinders the ability to assess scalability, fault tolerance, and adaptability under stress. }

Deep learning and AI-based methods, though promising, add layers of computational and interpretive complexity that are rarely stress-tested under constrained edge resources. The absence of standardized benchmarks, dynamic workload orchestration, and thorough evaluations across diverse scenarios remains a critical bottleneck. To unlock the full potential of edge-enabled robotics, future research must prioritize robust, scalable, and generalizable experimental frameworks—grounded in real-world constraints—that can systematically address the challenges of interoperability, resilience, communication, and continuous adaptation across multi-robot and multi-environment systems.

\section{Challenges for Future Research}
\label{sec:challenges}
Computational offloading for robots offers a multitude of benefits for ground and aerial robots in real-world deployments. It allows for leveraging the computational and storage facilities available at the remote resources to perform latency-sensitive and computationally-intensive operations, \textcolor{black}{particularly algorithms involving} deep learning or computer vision, and it also allows for maintaining communication between its distributed platform. 

However, network conditions associated with such frameworks \textcolor{black}{introduce implementation challenges related to} delay and energy consumption. Offloading is a critical step in increasing the quality of experience for robots and facilitating the execution of applications with low latency requirements. Undoubtedly, significant efforts were made to increase the robot's performance via computational offloading on edge-based systems. Several challenges, however, remain open study areas for further investigation.
Some of these problems are discussed more below.

\subsection{Security}
Security \cite{JahanF20} is a key performance parameter in all networked robotic applications. Communication and data transmission between robots and distant computing resources can lead to security concerns such as modification of data, denial of service, and information leakage \cite{1613648}. Unfortunately, in the studies mentioned above, security was not considered a primary optimization objective. Like other objectives stated, the security aspect should be incorporated as a key decision-making parameter to mitigate risks associated with data vulnerabilities. 

\subsection{Context Awareness} 
\textcolor{black}{As evident from Table} ~\ref{tab2}, context awareness has received limited attention in the design of offloading mechanisms, despite improvements in related works. Given that we are discussing dynamic computational offloading, context-awareness is a critical aspect that warrants deeper exploration.
It should inform every stage of the offloading process. For example, the program partitioning stage must be aware of the application's resource requirements to offload. The offloading decision should therefore be informed by the context information, such as distant server availability, network quality, and device workload, at all times \cite{Chaari2022}. 

\subsection{Handover Mechanism}
During the crossover between multiple edge devices, there is a huge possibility of data loss. While the task finishes itself on one edge device and switches to another, the robots either go into an idle state or halt their execution, which can introduce latency and can also become costly and a safety hazard while performing tasks like navigation. All dynamic computational offloading schemes must be designed to be equipped with an efficient handover mechanism to avoid computational cost, latency, or task failures.

\subsection{Network failure}
Connectivity loss is another major concern that can happen during offloading, where the robots end up losing connection with the remote resources and keep waiting for the communication to resume, when it could be a complete network failure or a temporary connectivity loss \cite{parasuraman2013spatial,parasuraman2023rapid}. Accordingly, it's beneficial if the robots take charge of the execution by performing the tasks locally, which should work ideally, but can be a serious problem in performing algorithms that have been split among resources to save on energy consumption and execution latency. 

\subsection{Fault tolerance}
The dependence on network circumstances is one of the most crucial issues in the computation of offloading for mobile robots. This might result in a failure of the offloading process, either due to a weak connection between robots and remote resources or an issue with the remote edge server.
Because mobile robots are intended to be used in critical scenarios like search and rescue missions, it is critical to avoid these problems and ensure the application's proper execution. Other alternatives must be devised whereby the robots can switch to local execution or offload their tasks to other emergency backup remote resources to avoid any delay in task execution. 

\subsection{Heterogeneity}
Although heterogeneity is a desirable feature in the implementation of the edge computing domain, its integration with robots can cause major interoperability concerns, which can be a hurdle in the effective deployment of an edge computing framework in a heterogeneous or otherwise multirobot system.

All such failures that might occur during task offloading of computational intensive tasks lead to either increased energy consumption or increasing latency time for task execution or data transmission. An efficient offloading strategy must take into account all such factors to make informed decisions or provide remedial solutions as part of the recovery mechanism; therefore, it is suggested that fault tolerance and recovery strategies must be an essential part of the offloading decision-making process. 

\section{Conclusion} 
\label{sec:conclusion}
\textcolor{black}{This study presents a comprehensive review of the current state-of-the-art in the edge robotics domain. While it offers a broad overview, it is not exhaustive and can be extended further, particularly by incorporating a detailed classification of the literature based on Fog computing technologies. The findings highlight that although many studies have leveraged edge platforms to address the computational demands of robotic applications, the field remains in its early stages. There is significant potential to pursue overlapping objectives and address persistent challenges related to the practical implementation of edge computing in robotics. These limitations—such as efficient resource management, scalable system architectures, robust failure recovery mechanisms, and strong security provisions—can be overcome through well-designed solutions tailored to the unique requirements of edge robotic systems.}

\bibliography{survey}
\bibliographystyle{IEEEtran}
\end{document}